\documentclass{article}
\usepackage{spconf,amsmath,amssymb,graphicx,hyperref}

\usepackage{algorithm}
\usepackage{algpseudocode}
\usepackage{multirow} 
\usepackage{booktabs} 
\usepackage{float}
\usepackage[font=small,labelfont=bf]{caption}  
\title{An Adaptive Edge-Guided Dual-Network Framework for Fast QR Code Motion Deblurring}
\name{Jianping Li$^{\star,\S}$, Dongyang Guo$^{\dagger}$, Wenjie Li$^{\ddagger}$, Wei Zhao$^{\ddagger}$ \thanks{Corresponding author: Wei Zhao.}}

\address{$^\star$ Guangdong Provincial Key Lab of Robotics and Intelligent Systems, \\
         Shenzhen Institute of Advanced Technology, Chinese Academy of Sciences, Shenzhen, China \\
         $^\S$ University of Chinese Academy of Sciences, Beijing, China \\
         $^\dagger$ School of Nano Science and Technology, University of Science and Technology of China, Suzhou, China \\
         $^\ddagger$  Shenzhen University of Advanced Technology, Shenzhen, China
         }

\begin{document}
%
\maketitle
\begin{abstract}
Unlike general image deblurring that prioritizes perceptual quality, QR code deblurring focuses on ensuring successful decoding. QR codes are characterized by highly structured patterns with sharp edges, a robust prior for restoration. Yet existing deep learning methods rarely exploit these priors explicitly. To address this gap, we propose the Edge-Guided Attention Block (EGAB), which embeds explicit edge priors into a Transformer architecture. Based on EGAB, we develop Edge-Guided Restormer (EG-Restormer), an effective network that significantly boosts the decoding rate of severely blurred QR codes. For mildly blurred inputs, we design the Lightweight and Efficient Network (LENet) for fast deblurring. We further integrate these two networks into an Adaptive Dual-network (ADNet), which dynamically selects the suitable network based on input blur severity, making it ideal for resource-constrained mobile devices. Extensive experiments show that our EG-Restormer and ADNet achieve state-of-the-art performance with a competitive speed. Project page: \url{https://github.com/leejianping/ADNet}

\end{abstract}

\begin{figure}[htbp]
\centering
\begin{minipage}[b]{1.0\linewidth}
  \centering
  \centerline{\includegraphics[width=8.5cm]{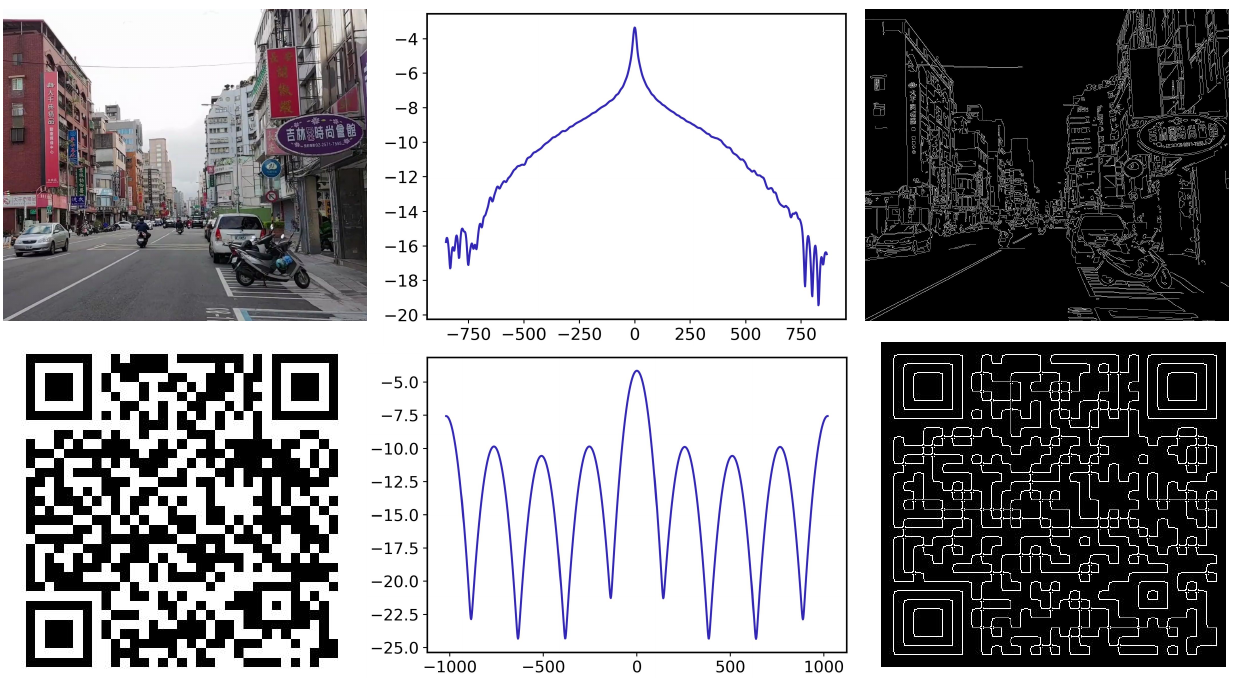}}
\end{minipage}

\begin{minipage}[b]{0.32\linewidth}
  \centering
   \fontsize{9pt}{10.8pt}\selectfont (a) sharp image
\end{minipage}
\hfill
\begin{minipage}[b]{0.32\linewidth}
  \centering
  \fontsize{9pt}{10.8pt}\selectfont  (b)log-scale gradient histograms
\end{minipage}
\hfill
\begin{minipage}[b]{0.32\linewidth}
  \centering
  \fontsize{9pt}{10.8pt}\selectfont (c) gradient magnitude
\end{minipage}
\caption{ Comparison of gradient statistics: natural image vs QR code.}

\label{fig:sig}
\end{figure}

\section{Introduction}
\label{sec:intro}

Image deblurring is a fundamental task in low-level computer vision aimed at recovering high-quality images from blurry inputs. Motion deblurring is an important sub-task that removes blur caused by camera shake or object movement~\cite{zhang, cho}. Even minor motion blur from camera shake can render QR codes unreadable, motivating extensive research into motion deblurring methods for QR codes~\cite{soros, lv, alam2025}. Early works include Van et al.~\cite{van2015}, who proposed a regularization method using corner features for point spread function (PSF) estimation, and Pan et al.~\cite{2019kl}, who exploited barcode symbology properties for kernel estimation via Kullback-Leibler divergence. However, these traditional methods often struggle with complex, non-uniform motion blur in practical scenarios, and typically fail to restore severely blurred QR codes.

To overcome these limitations, deep learning has become the dominant paradigm for image deblurring, capable of learning effective features from large-scale datasets. Representative architectures include U-shaped networks~\cite{general, sfnet}, generative adversarial networks~\cite{deblurgan}, self-attention mechanisms~\cite{restormer, image}, and nonlinear activation-free networks~\cite{simple}. While these methods excel in general image deblurring, their direct application to domain-specific tasks, such as QR code or text deblurring, typically result in suboptimal performance. This is mainly due to the unique structural properties of QR codes, which are two-dimensional barcodes composed of a grid of black and white modules with sharp edges and precise spatial arrangements. This inherent structure leads to a unique statistical property: as shown in Fig.~\ref{fig:sig}(b), the gradient histogram of a QR code image differs significantly from that of a natural image. Moreover, the strong gradients in QR codes exhibit a highly regular spatial distribution (Fig.~\ref{fig:sig}(c)) due to their grid-aligned arrangements. 

\begin{figure*}[htbp]
  \centering
   
  \includegraphics[width=\linewidth]{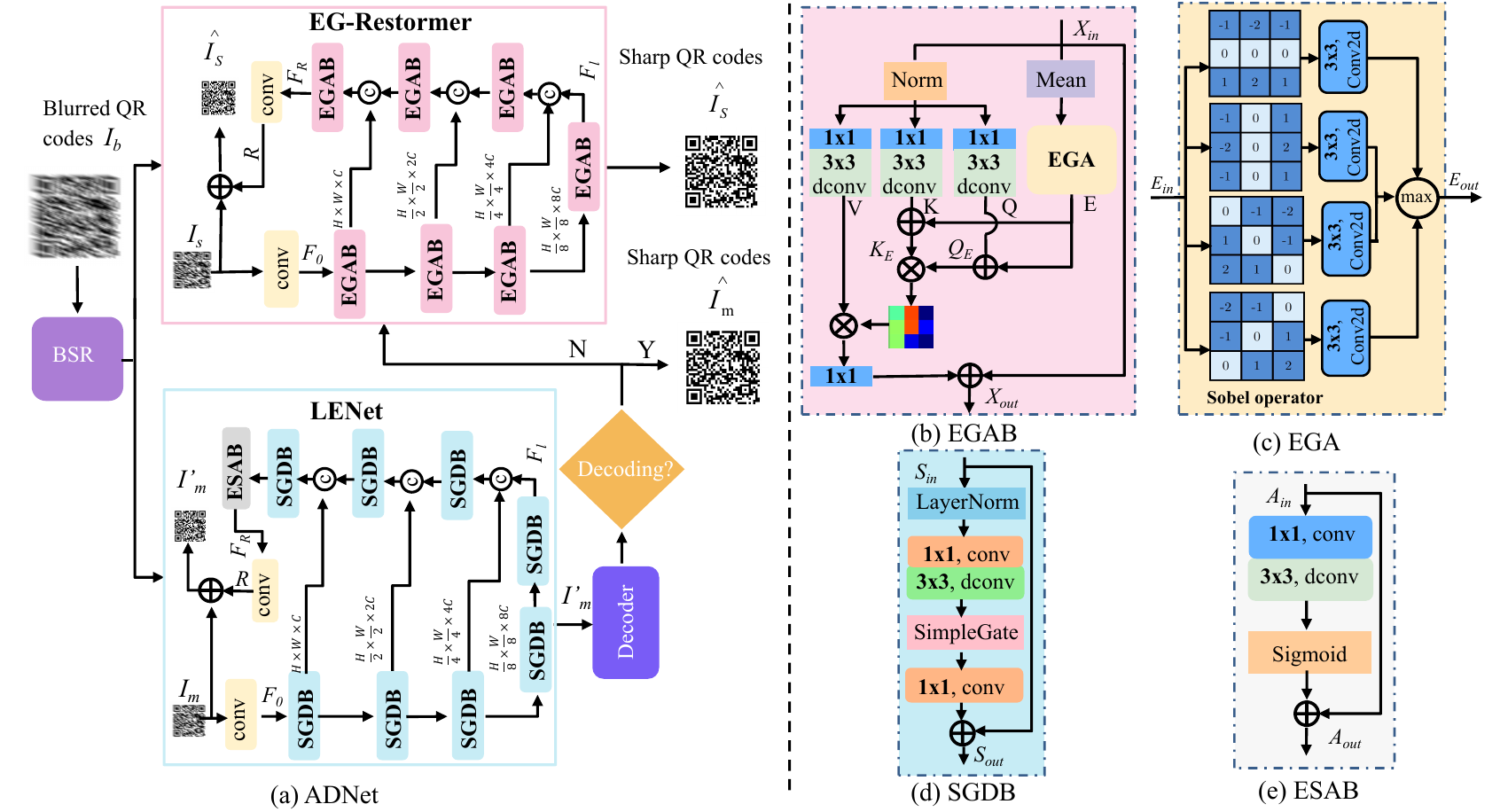}
  \caption{The overall architecture of the proposed ADNet framework. (a) The ADNet consists of a powerful EG-Restormer and the LENet. (b) The EGAB uses the EGA module to inject an explicit edge feature $E$ into the attention mechanism. (c) The structure of the EGA module. (d) The architecture of the SGDB. (e) The structure of the edge sharpening attention block (ESAB).}
\label{fig:frame}

\end{figure*} 

Consequently, developing domain-specific models for QR code deblurring is essential. Recently, Pu et al.~\cite{puquick} presented a dual CNN framework for reversing degradation in 2D barcodes. Li et al.~\cite{limotion} proposed a deep convolutional network with an adaptive thresholding technique to handle motion blur and uneven illumination. Additionally, Dong et al.~\cite{donggan} and Wang et al.~\cite{motiongan} developed GAN-based methods for restoring severely motion-blurred QR codes. Although these works~\cite{puquick,limotion,donggan,motiongan} adapt general deblurring models to the QR code domain,  they do not explicitly incorporate the domain-specific priors inherent to QR codes.

Despite recent advances in QR code motion deblurring, two critical challenges persist: (1) Existing learning-based methods rely on implicit feature learning and fail to explicitly model the inherent edge priors that are critical for successful QR code decoding; (2) These approaches often overlook inference efficiency, a vital requirement for deployment on resource-constrained mobile devices.

To tackle these issues, we propose EG-Restormer—an efficient Transformer-based model that embeds edge priors explicitly via our Edge-Guided Attention Block (EGAB). We further develop an Adaptive Dual-Network (ADNet) framework that balances performance and efficiency by dynamically routing inputs to either a large network (EG-Restormer) or a lightweight one (LENet) based on the severity of input blur. Extensive evaluations show our method achieves state-of-the-art performance in both restoration quality and decoding rate, while maintaining competitive execution speed.

\section{PROPOSED METHOD}
\label{sec:method}

\subsection{Overall Architecture}
\label{ssec:overall}

As shown in Fig.~\ref{fig:frame}(a),  ADNet integrates two pre-trained QR code deblurring models: EG-Restormer and LENet. Both models adopt a U-shaped architecture; however, LENet—built with the Simple Gate Depthwise Convolution Block (SGDB) without transformers—is lighter and faster than EG-Restormer, which uses EGAB. For a given blurred input QR code $I_{b} \in \mathbb{R}^{H \times W \times 3}$ (height H, width W), a Blur Severity-based Routing (BSR) unit classifies $I_b$ into two categories based on blur severity: severely blurred inputs $I_s$ are directed to EG-Restormer for a sharpened output, and mildly blurred inputs $I_m$ are directed to LENet for fast deblurring. If the output of LENet is successfully decoded, it becomes the final restored QR code $\hat{I}_m$; otherwise, $I_m$ is re-routed to EG-Restormer for further refinement. 

In the EG-Restormer's U-shaped design, the input $I_{s}$ is passed through a convolution layer to extract low-level features ${F_{0}}$. These features are then processed by a four-layer encoder that progressively downsamples spatial dimensions, while expanding the channels for deep feature extraction. The decoder then iteratively upscales the bottleneck feature ${F_{l}}$ to high-resolution features ${F_{r}}$. Finally, the output features are convolved to produce a residual QR code $R$ added to the input for restoration. Both the encoder and decoder include 4 EGABs to guide the restoration of edge information. Similarly, LENet processes mildly blurred QR codes using a four-level SGDB-based encoder-decoder. Its decoder output is fed to an Edge Sharpening Attention Block (ESAB), which includes a 3x3 depthwise convolution for spatial context refinement and edge enhancement (Fig.~\ref{fig:frame}(e)).

\subsection{Blur Severity-based Routing Unit}
\label{ssec:bsr}
The core component of our ADNet is the BSR unit, as outlined in Algorithm~\ref{alg} . It categorizes input QR codes as mildly or severely blurred using a pre-calibrated threshold $\tau$ on Laplacian variance (LV), a sharpness metric where a higher LV indicates less blur. This threshold is derived from LENet's performance on a test dataset. Specifically, $\tau$ is calculated as the midpoint between the minimum LV of the blurriest decodable restored QR codes and the maximum LV of the sharpest non-decodable ones:

\begin{equation}
\tau = \frac{\min(\mathcal{L}_{\text{decodable}}) + \max(\mathcal{L}_{\text{non-decodable}})}{2}
\end{equation}

where $\mathcal{L}_{\text{decodable}}$  and $\mathcal{L}_{\text{non-decodable}}$ denote the set of LV scores for QR codes that are  decodable and  non-decodable after LENet restoration, respectively.
\begin{algorithm}[H]
\caption{Blur Severity-based Routing Strategy}
\label{alg}
\begin{algorithmic}[1] 

\State \textbf{Input:}
 $I_{b}$: A blurred QR code, 
\State \quad $M_{E}$: The large network (EG-Restormer).
\State \quad $M_{L}$: The lightweight network (LENet).
\State \quad $\tau$: A pre-calibrated blur severity threshold.
\State \quad $\mathcal{M}(\cdot)$: A blur metric function (Laplacian Variance).

\Function{BlurRouter}{$I_{b}, M_{E}, M_{L}, \tau, \mathcal{M}$}
    \State $v \gets \mathcal{M}(I_{b})$ \Comment{Calculate the blur severity score}
    
    \If{$v > \tau$} \Comment{Case 1: Mild blur detected}
        \State $M_L \gets I_{b}$ \Comment{Route to LENet}
    \Else \Comment{Case 2: Severe blur detected}
        \State $M_E \gets I_{b}$ \Comment{Route to EG-Restormer}
    \EndIf
    
\EndFunction
\end{algorithmic}
\end{algorithm}

\subsection{Edge-Guided Attention Block}
\label{ssec:egab}
Edge sharpening is crucial for successful QR code decoding. Conventional learning-based methods often rely on implicit feature extraction, overlooking explicit modeling of edge priors—such as locations and gradients—that are vital for decoding but pose quadratic complexity in attention mechanisms. To address this, we propose the EGAB, as illustrated in Fig.~\ref{fig:frame}(b), which embeds QR code edge priors by modulating the query and key in multi-Dconv head transposed attention (MDTA)~\cite{restormer}. 

The EGAB generates an edge-map $E$ using the proposed Edge-Guided Attention (EGA) module to capture explicit edge features, while producing query (Q), key (K) and value (V) matrices from the input features for implicit spatial representation. The query and key are then modulated by the edge map to yield $Q_E$ and  $K_E$, followed by computation of an edge-guided transposed-attention map through dot-product between the reshaped $Q_E$ and $K_E$. The process is formulated as:

\begin{equation}
\begin{aligned}
&X_{out} = Attention({{\hat{Q}}_E}, {{\hat{K}}_E}, {{\hat{V}}}) + X_{in},\\
&Q_E = Q (1 + W_E \times E),\\
&K_E = K (1 + W_E \times E),\\
&E = EGA(Mean(X_{in})) 
\end{aligned}
\end{equation}

Where $X_{in}$ and $X_{out}$ are input and output features; $W_E$ weights the edge map, and matrices $\hat{Q}_E$, $\hat{K}_E$, $\hat{Q}$, $\hat{K}$, and $\hat{V}$ are derived by reshaping input tensors from their original dimensions $\mathbb{R}^{H \times W \times C}$. 

As shown in Fig.~\ref{fig:frame}(c), the EGA is designed to capture multi-directional edge features. The single-channel input feature $E_{in}$ is processed by four parallel Sobel operators to detect edges in four orientations (horizontal, vertical, $45^{\circ}$ and $135^{\circ}$). These directional edge features are filtered through fixed-weight convolutions to generate four edge feature maps. A max operation selects the dominant map as the final output. 


\subsection{Simple Gate Depthwise Convolution Block}
\label{ssec:sgdb}

Running a high-capacity model on all inputs is inefficient for resource-constrained mobile devices, especially since the error correction built into QR codes tolerates mild blur. Inspired by MobileNetV2~\cite{mobile} and NAFNet~\cite{simple}, we propose the Simple Gate Depthwise Convolution Block (SGDB) for efficient feature extraction and refinement.
As shown in Fig.~\ref{fig:frame}(d), the input feature $S_{in}$ is processed through: layer normalization (LN) , 1x1 convolution (Conv) for channel expansion, 3x3 depthwise convolution (Conv2d) for spatial context aggregation, SimpleGate (SG)~\cite{simple}, and another 1x1 convolution for channel contraction, with a weighted residual connection producing refined output $S_{out}$.




\section{Experiments}
\label{sec:experiment}

\subsection{Datasets and Metrics}
\label{ssec:data}

\begin{table*}[t] 
\centering
\caption{Comparison of different deblurring methods and training strategies on the QR code test set.}
\label{tab:main_comparison}
\begin{tabular}{@{}lccccccccc@{}} 
\toprule
\multirow{2}{*}{\textbf{Methods}} & \multicolumn{3}{c}{\textbf{GoPro}} & \multicolumn{3}{c}{\textbf{GoPro + QRData}} & \multirow{2}{*}{\textbf{Params.(M)}} & \multirow{2}{*}{\textbf{Flops(G)}} & \multirow{2}{*}{\textbf{Avg\_time(s)}} \\
\cmidrule(r){2-4} \cmidrule(l){5-7} 
& \textbf{DR(\%)} & \textbf{PSNR} & \textbf{SSIM} & \textbf{DR(\%)} & \textbf{PSNR} & \textbf{SSIM} & & & \\
\midrule
LENet (Ours) & 32.67 & 9.78 & 0.421 & 49.33 & 11.37 & 0.572 & \ \textbf{0.28} & \textbf{3.87} & \textbf{0.23} \\
NAFNet-32 \cite{simple} & 40.67 & 10.20 & 0.449 & 81.33 & 15.56 & 0.695 & 17.11 & 64.44 & 0.794 \\
Restormer \cite{restormer} & 50.00 & 10.60 & 0.496 & 88.67 & \textbf{18.15} & \textbf{0.766} & 26.13 & 564.96 & 0.838 \\
EG-Restormer (Ours) & \textbf{58.67} & \textbf{10.99} & \textbf{0.582} & \textbf{90.00} & 14.88 & 0.666 & 26.13 & 565.14 & 0.910 \\
\bottomrule
\end{tabular}
\end{table*}

We created a synthetic QR code Dataset (QRData), containing 1,822 images blurred with non-uniform motion kernels for training and evaluation. Our models are trained using a two-stage transfer learning strategy: pre-training on the GoPro~\cite{gopro} dataset (2,103 training images, 1,111 test images), followed by fine-tuning on QRData (1,521 training images, 301 testing images). For performance assessment, we construct a test set of 150 motion blurred QR codes, including 100 synthetic and 50 real-world captures under realistic conditions. Evaluation metrics include Peak Signal-to-Noise Ratio (PSNR), Structural SIMilarity (SSIM), and the Decoding  

\begin{figure}[htb]

\begin{minipage}[b]{1.0\linewidth}
  \centering
  \centerline{\includegraphics[width=8.5cm]{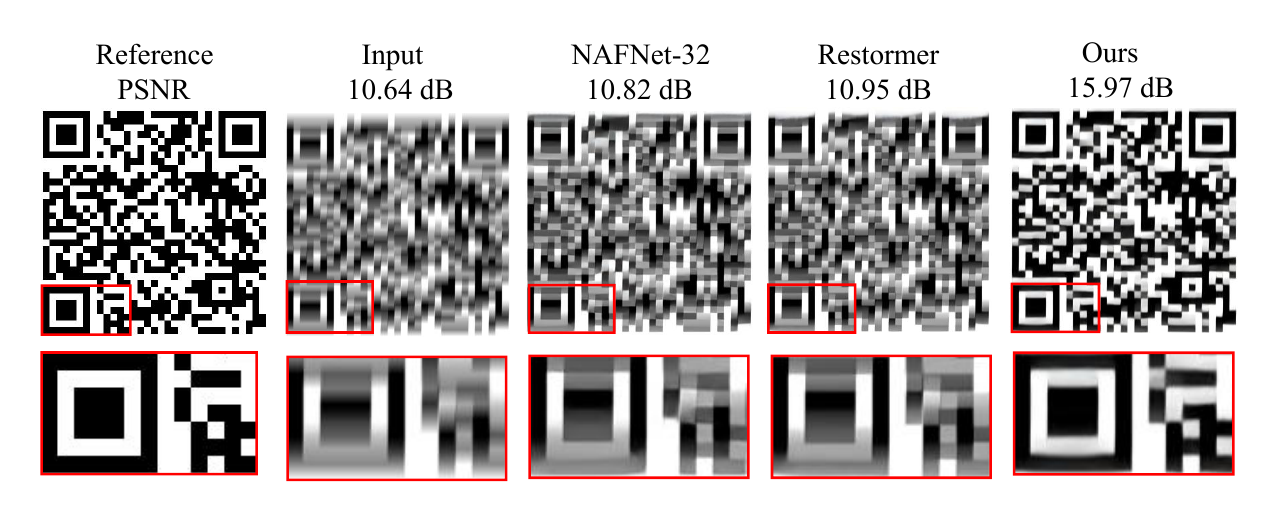}}
\end{minipage}
\caption{Qualitative comparison of QR code deblurring on the test set.}
\label{fig:res}
\end{figure}

\noindent Rate (DR), defined as the percentage of successfully decoded QR codes. Model complexity is evaluated via parameter count and FLOPs (for a 512 $\times$ 512 image), while efficiency is measured by Average inference time (Avg$\_$time) per image in seconds.

\subsection{Implementation Details}
\label{ssec:implement}

EG-Restormer is configured with [4, 6, 6, 8] Transformer blocks and [1, 2, 4, 8] attention heads across four levels, using 48 base channels~\cite{restormer}. All models are trained using the AdamW optimizer (${\beta}_{1}$=0.9, ${\beta}_{2}$=0.999, weight decay = 0.0001). The initial learning rate is $3e^{-4}$  and reduced to $1e^{-6}$ via cosine annealing. EG-Restormer employs progressive training for 400K iterations with (patch size, batch size) pairs:[($128^2$, 64), ($160^2$, 40), ($192^2$, 32), ($256^2$, 16), ($320^2$, 8), ($384^2$, 8)]. LENet is trained for 1,000K iterations with a fixed 256×256 patch size and batch size of 8. For data augmentation, we use flipping, rotation, and shuffling. For ADNet, the Zbar library was used for all decoding attempts.

\subsection{Results and Discussion}
\label{ssec:restult}

We compare EG-Restormer against state-of-the-art image deblurring methods, NAFNet-32~\cite{simple} and Restormer~\cite{restormer}. Quantitative results for two training strategies are presented in Table~\ref{tab:main_comparison}. When pre-training only on the GoPro,  EG-Restormer outperformed Restormer by 0.39 dB in PSNR and 8.67$\%$ in DR. After fine-tuning on QRData, our model maintains its superiority,
\begin{table}[h!]
\centering
\caption{Ablation studies for blur severity-based routing unit.} 
\label{tab:eff} 
\begin{tabular}{lcc}
\toprule 
\textbf{Methods} & \textbf{Avg\_time(s)} & \textbf{DR(\%)} \\
\midrule 
Restormer (baseline)  & 0.838 & \textbf{88.67} \\
EG-Restormer   & 0.910 & \textbf{90} \\
LENet  & \textbf{0.28} & 49.33 \\
ADNet (random) & 0.457 & 72 \\
ADNet (ours)  & 0.737 & \textbf{90} \\
\bottomrule 
\end{tabular}
\end{table}
\noindent achieving a 1.33$\%$ higher DR than the baseline. These results demonstrate that our EG-Restormer achieves state-of-the-art performance. Although Restormer achieves higher PSNR/SSIM after fine-tuning, our EG-Restormer's superior DR emphasizes that explicit edge guidance is more critical for decodability than pixel-wise accuracy. Furthermore, Fig.~\ref{fig:res} shows that the proposed EG-Restormer is able to restore severely motion blurred QR codes, outperforming two other methods in visual quality, validating the effectiveness of incorporating explicit edge priors.
 
\textbf{Ablation Study} As shown in Table~\ref{tab:eff}, we conducted an ablation experiment on our blur severity-based routing strategy by comparing it with three baseline methods: exclusive use of EG-Restormer, exclusive use of LENet, and random selection (ADNet-Random). The results clearly demonstrate the benefits of our approach. ADNet matches the 90\% decoding rate (DR) of the powerful EG-Restormer while reducing the Avg$\_$time by 19\% (from 0.910s to 0.737s). Compared to random routing, ADNet boosts the DR by 18\%. This outcome validates that our blur severity-based routing strategy effectively balances performance and efficiency by intelligently assigning tasks to the appropriate network.

\section{CONCLUSION}
\label{sec:conclusion}
In this paper, we proposed a novel approach for domain-specific deblurring by directly integrating edge priors into a Transformer-based network. Our proposed Edge-Guided Attention Block (EGAB) enables the network to focus on restoring the sharp edges crucial for QR code recognition. Furthermore, our adaptive dual-network (ADNet) framework intelligently combines a lightweight network with a powerful network, significantly improving inference efficiency without sacrificing performance. Comprehensive experiments demonstrate that our method achieves state-of-the-art performance in QR code restoration, characterized by superior decoding rates, competitive image quality metrics, and higher inference speed. This work provides a promising direction for adapting general vision models to specialized tasks with strong structural priors, potentially benefiting other areas like text image deblurring or face image deblurring.

\vfill\pagebreak

\bibliographystyle{IEEEbib}
\bibliography{adnet}

\end{document}